\documentclass[letterpaper]{article} 
\usepackage[]{aaai2027}  
\usepackage[hyphens]{url}  
\usepackage{graphicx} 
\usepackage{natbib}  
\usepackage{caption} 
\usepackage{algorithm}
\usepackage{algorithmic}
\usepackage{amsmath} 
\usepackage{amssymb}
\usepackage{multirow} 
\usepackage{newfloat}
\usepackage{listings}
\DeclareCaptionStyle{ruled}{labelfont=normalfont,labelsep=colon,strut=off} 
\floatstyle{ruled}
\newfloat{listing}{tb}{lst}{}
\floatname{listing}{Listing}

\usepackage{booktabs}

\title{SPEAR: A Spectral-Disentangled MoE Neural Operator with Knowledge-Guided Expert Aggregation for Large-Scale PDE Pretraining}

 \author{
 	Dengdi Sun$^1$, Xiaoya Zhou$^1$, Xiao Wang$^2$*, Wanli Lyu$^2$, Jin Tang$^2$, Bin Luo$^2$
 }
\affiliations{
1. School of Artificial Intelligence, Anhui University, Hefei 230601, China \\ 
2. School of Computer Science and Technology, Anhui University, Hefei 230601, China \\ 
}

\begin{document}

\maketitle

\begin{abstract}
Large-scale pre-training has improved the generalization of neural operators across diverse PDEs. However, existing PDE foundation models still struggle with heterogeneous dynamics, where shared representations may cause knowledge interference, while mixture-of-experts (MoE) architectures suffer from increasing expert redundancy. We propose SPEAR, a spectral-disentangled MoE neural operator with knowledge-guided expert aggregation for large-scale PDE pre-training. SPEAR decouples latent features into low- and high-frequency components, enabling shared modeling of transferable dynamics and specialized learning of PDE-specific patterns. To address expert redundancy, we design a knowledge-guided expert aggregation strategy that measures expert similarity from dataset-specific learned knowledge and routing preferences, enabling the identification and consolidation of similar experts. Experiments on twelve PDE datasets and multiple downstream benchmarks demonstrate superior performance in pre-training, fine-tuning, and transfer learning. Furthermore, our aggregation strategy reduces the number of experts by 50\% while maintaining or improving prediction accuracy, achieving a balance between model efficiency and generalization for PDE foundation models.
\end{abstract}

\section{Introduction}
Neural operators have emerged as a compelling alternative for solving PDEs by learning mappings between infinite-dimensional function spaces, achieving orders-of-magnitude speedups at inference time. However, existing neural operators are typically trained under single-task settings—specific PDE families with fixed parameter ranges—resulting in limited generalization to unseen equation types, parameter regimes, or boundary conditions. Inspired by large language models \cite{vaswani2017attention} and vision foundation models \cite{radford2021learning}, recent efforts have introduced large-scale pre-training into the PDE domain to learn transferable physical representations. While pre-training enhances generalization, different PDEs exhibit substantially different underlying dynamics, and modeling multiple physical processes within a single shared parameter space may lead to knowledge conflicts and negative transfer.

\begin{figure}
    \centering
    \includegraphics[width=1\linewidth]{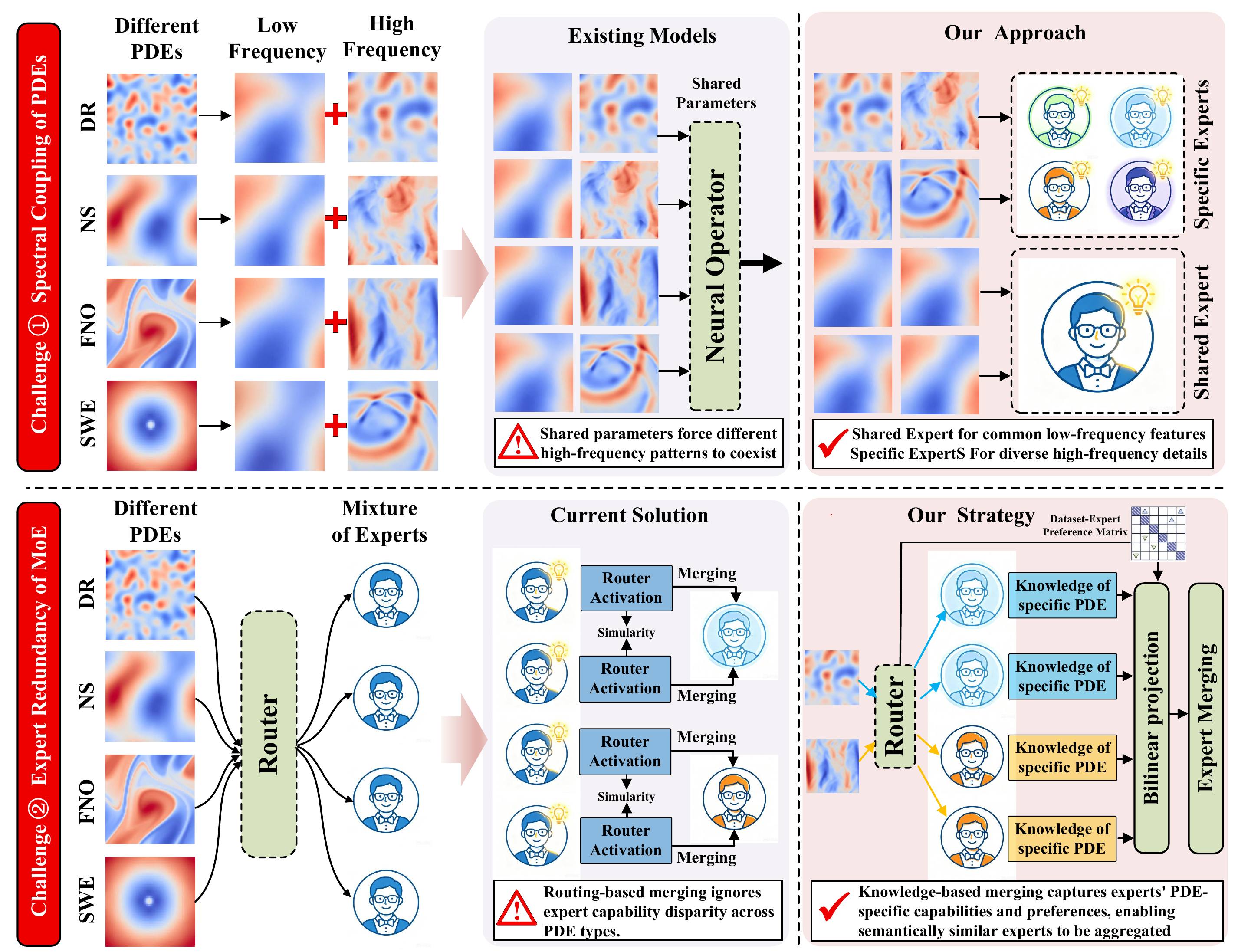}
    \caption{Motivation of SPEAR. The upper shows the advantage of frequency decoupling for separating shared and PDE-specific dynamics. The lower highlights knowledge-guided expert merging over routing-based merging.}
    \label{fig:motivation}
\end{figure}

To scale up modeling capacity while maintaining inference efficiency, recent work has introduced Mixture-of-Experts (MoE) architectures \cite{shazeer2017outrageously} into neural operator design \cite{wang2026mixture, sun2026nestor}. MoEs employ dynamic routing mechanisms that activate only a subset of expert subnetworks for each input, enabling massive parameter scaling without proportionally increasing per-sample computation. 
This paradigm shows promise in handling PDE heterogeneity: different experts can specialize in different equation types, physical regimes, or numerical patterns. However, MoE-based neural operators still face the following two critical challenges. As shown in Figure~\ref{fig:motivation}, we summarize the motivation and overview of SPEAR.

\textbf{Challenge 1: Spectral Coupling of PDEs}. Different PDE families differ drastically in their physical behaviors. For example, elliptic equations describe steady-state equilibrium, with high-frequency features manifesting as smooth local extrema; hyperbolic equations describe wave propagation, with high-frequency features manifesting as sharp discontinuities; parabolic equations describe dissipative processes, with high-frequency features manifesting as decaying spikes. These differences are most pronounced in high-frequency local patterns, while low-frequency dynamics exhibit broad similarity across equation types. However, Fourier-based neural operators such as AFNO \cite{guibas2021adaptive} couple all frequency components within shared parameters, forcing morphologically distinct high-frequency patterns to coexist and causing severe gradient interference. Existing remedies, including multi-scale operators \cite{LIU2024112944} and frequency enhancement modules \cite{10.24963/ijcai.2025/818}, have yet to break this spectral coupling constraint. Therefore, a mechanism that can separate shared dynamics from specific patterns according to frequency characteristics is urgently needed.

\textbf{Challenge 2: Expert Redundancy of MoE}. While MoE architectures enhance model capacity through specialization, this comes at a cost: as model scale grows, parameter redundancy proliferates, with many experts learning functionally similar representations while inference overhead increases significantly. Existing expert merging methods \cite{zhang2026expert, li2026submoe, li2024merge, chen2024retraining} primarily consolidate similar experts based on routing statistics or expert outputs, rather than the task-specific knowledge semantics that experts truly acquire. This "behavior-based" rather than "knowledge-based" merging strategy lacks physical grounding. In PDE learning, this is particularly problematic: two experts that handle similar physical patterns may never be activated together due to different data distributions, and would thus be judged as "dissimilar" and left unmerged.

It is worth noting that these two challenges are not independent. A key fact connects them: low-frequency dynamics are broadly transferable across PDE types and can be effectively captured by a shared backbone, while high-frequency patterns are PDE-specific and demand expert specialization. This observation naturally suggests a two-stage strategy: first, decouple frequencies to separate sharable low-frequency knowledge from PDE-specific high-frequency patterns, assigning the latter to a dedicated MoE module for expert-level specialization; then, consolidate the high-frequency experts to eliminate redundancy, ensuring that each preserved expert captures truly distinct physical knowledge. In other words: decouple to specialize, then aggregate to refine.


Building on this insight, in this paper, we present a \textbf{SP}ectral-disentangled MoE neural operator with knowledge-guided \textbf{E}xpert \textbf{A}gg\textbf{R}egation (\textbf{SPEAR}). Our framework introduces two tightly coupled innovations. \textbf{For challenge 1}, we propose a spectral-disentangled architecture that explicitly partitions the frequency spectrum. The low-frequency branch employs AFNO to capture global evolution patterns, while the high-frequency branch extracts fine-grained local structures via high-pass filtering and models them through an MoE module composed of shared convolutional backbones and LoRA adapters. Dynamic routing based on high-frequency statistics activates a part of experts to efficiently learn local dynamics across PDEs. \textbf{For challenge 2}, we design a knowledge-guided expert aggregation strategy that compresses and reorganizes functionally similar experts by incorporating physical correlations among PDE tasks and expert preference information. Rather than routing statistics, expert similarity is measured by the learned parameters for each PDE dataset. Then, a dataset-level knowledge matrix is constructed and fused via bilinear projection, yielding semantic similarities for expert merging. Our main contributions are summarized as follows:

1) We design a spectral-disentangled architecture that explicitly partitions the frequency spectrum: a shared backbone captures low-frequency dynamics across all PDE types, while a dedicated MoE module handles PDE-specific high-frequency patterns.

2) We propose a knowledge-guided expert aggregation strategy that computes similarities among experts based on their acquired physical knowledge during training and routing preference information, and merges them via bilinear projection and hierarchical clustering.

3) Extensive experiments on large-scale PDE pretraining datasets show that our method learns transferable representations, achieving strong performance across cross-equation pretraining, fine-tuning, and generalization to unseen PDEs with superior parameter efficiency.

\section{Related Work}
\subsection{Neural Operators for PDE Solving}
Neural operators learn mappings between infinite-dimensional function spaces~\cite{lu2021learning,li2020fourier}. Various neural operators have been developed for PDE modeling. For example, DeepONet~\cite{lu2021learning} adopts a branch--trunk architecture, while the Fourier Neural Operator (FNO)~\cite{li2020fourier} models global dependencies in the Fourier domain. More recently, methods such as MPP~\cite{mccabe2023multiple}, DPOT~\cite{hao2024dpot}, Poseidon~\cite{herde2024poseidon}, OmniArch~\cite{chen2024omniarch}, and UniSolver~\cite{zhouunisolver} have explored pre-training on diverse PDE datasets, advancing PDE foundation models.
However, most existing methods adopt a unified architecture. As PDE datasets become larger and more diverse, differences in physical laws and dynamics may cause knowledge interference across heterogeneous tasks. Therefore, modeling diverse physical knowledge remains an important challenge in large-scale PDE pre-training.

\subsection{Mixture-of-Experts Models}
Mixture-of-Experts (MoE) expands model capacity through sparse expert activation and has been widely adopted in language and vision models. Representative works include Sparsely-Gated MoE~\cite{shazeer2017outrageously}, Switch Transformer~\cite{fedus2022switch}, and Mixtral~\cite{jiang2024mixtral}. By activating only a small subset of experts for each input, MoE increases model capacity with limited additional computation. Recently, MoE has also been introduced into neural operator models. Methods such as MoE-POT~\cite{wang2026mixture} and NESTOR~\cite{sun2026nestor} use specialized experts to capture diverse physical knowledge and alleviate interference across heterogeneous PDE tasks. However, some experts may learn similar physical representations, leading to knowledge redundancy and inefficient parameter utilization. Therefore, reducing redundant experts while preserving model performance remains an important challenge in MoE-based PDE modeling.

\subsection{Expert Merging}
Expert merging is an emerging compression technique that identifies experts with similar functions or representations and integrates them through parameter fusion. Representative methods include SMoE~\cite{li2024merge} and HC-SMoE~\cite{chen2024retraining}, which construct expert relationships from routing patterns or output features and then merge redundant experts. However, most existing methods lack explicit modeling of task semantics and domain knowledge, making it difficult to accurately characterize the knowledge learned by different experts.

In PDE modeling, MoE encourages functional specialization, while knowledge overlap may still exist among some experts. Incorporating physical knowledge into expert relationship modeling can therefore improve the identification of truly redundant experts and enable effective compression while preserving expert diversity, which is important for building efficient and scalable PDE foundation models.

\begin{figure}[t]
    \centering
    \includegraphics[width=\linewidth]{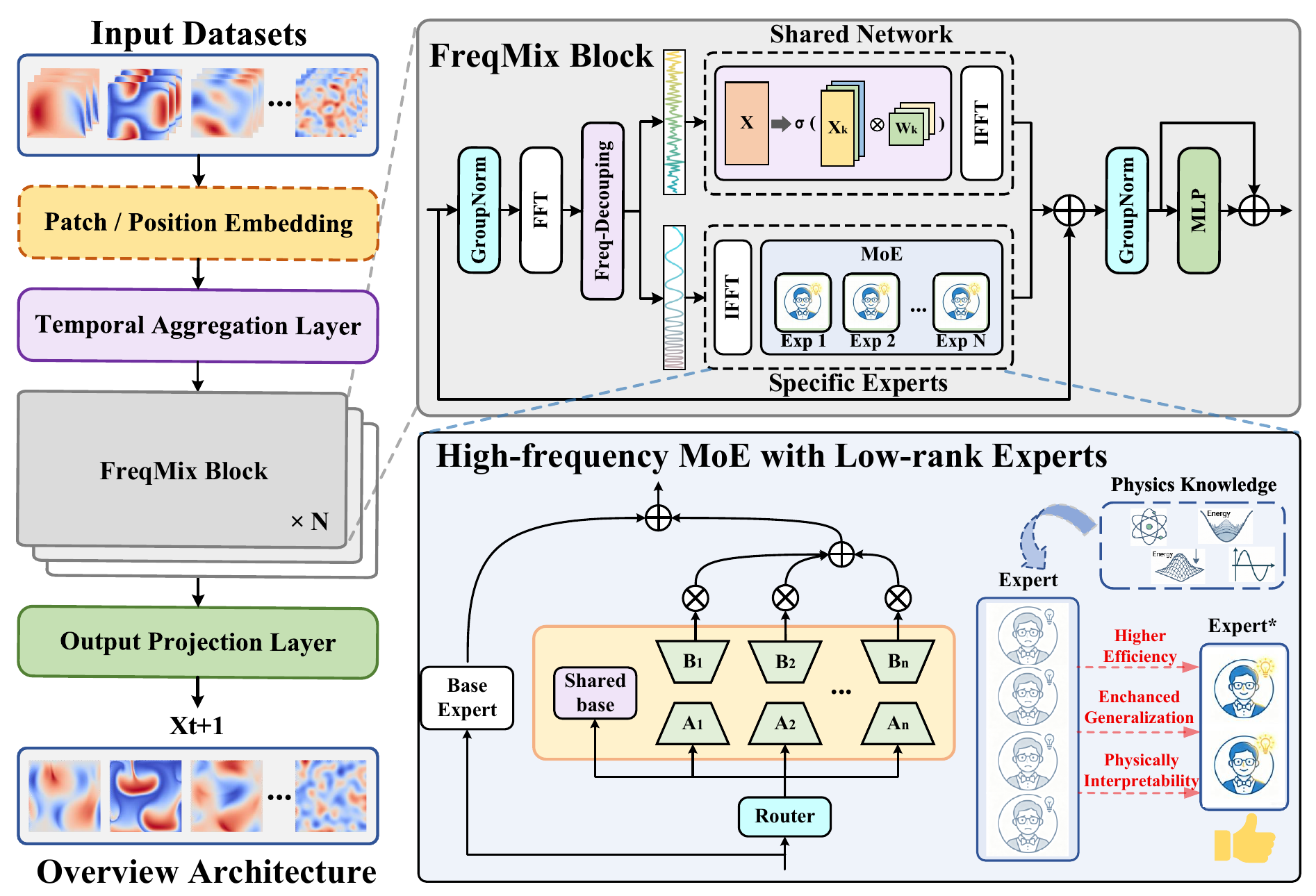}
    \caption{Overview architecture. The left panel shows the overall pipeline, the upper-right panel presents the FreqMix block, and the lower-right panel illustrates the high-frequency MoE module with low-rank experts.}
    \label{fig:架构图}
\end{figure}

\section{The Proposed Approach} 
We propose SPEAR, a spectral-disentangled MoE neural operator with knowledge-guided expert aggregation, as shown in Fig.~\ref{fig:架构图}. SPEAR decouples shared low-frequency dynamics from PDE-specific high-frequency patterns, where the MoE module enables expert specialization for local dynamics. Parameter-efficient design and knowledge-guided expert aggregation further reduce expert redundancy.

\subsection{Preliminaries}
We consider parameterized PDEs over a spatial domain $\Omega \subset \mathbb{R}^n$ and a time interval $[0,T]$:
\begin{equation}
\frac{\partial \mathbf{u}}{\partial t}
-
\mathcal{H}(\mathbf{u},\nabla \mathbf{u},\nabla^2\mathbf{u},\ldots;\theta)
=0,
\end{equation}
subject to
\begin{equation}
\begin{cases}
\mathbf{u}(\mathbf{x},0)=u_0(\mathbf{x}), & \mathbf{x}\in\Omega,\\
\mathcal{B}[\mathbf{u}](\mathbf{x},t)=g(\mathbf{x},t), &
(\mathbf{x},t)\in\partial\Omega\times[0,T],
\end{cases}
\end{equation}
where $u$ is the system state, $\mathcal{H}$ is the governing operator, $\theta$ denotes the physical parameters, and $\mathcal{B}$ is the boundary operator.
In the data-driven setting, a neural operator $\mathcal{T}_w$ approximates the temporal evolution:
\begin{equation}
u_{t+1}
=
\mathcal{T}_w
\left(
u_{t-K+1},\ldots,u_t
\right),
\end{equation}
where $w$ denotes the learnable parameters. Following DPOT~\cite{hao2024dpot}, small random perturbations are added to the input sequence during training to reduce error accumulation and improve long-term prediction robustness.

\subsection{Input Encoding and Time Aggregation}
Given an input PDE sequence
$\mathbf{x} \in \mathbb{R}^{B \times H \times W \times T \times C}$,
we concatenate it with the coordinate grid and reshape it into
$\widetilde{\mathbf{x}} \in \mathbb{R}^{(BT)\times(C+3)\times H\times W}$.
Following the patch embedding of ViT~\cite{dosovitskiy2020image}, the input is encoded as
\begin{equation}
\mathbf{X}
=
\operatorname{PatchEmbed}(\widetilde{\mathbf{x}}) + \mathbf{E}_{\mathrm{pos}}
\in
\mathbb{R}^{(BT)\times D\times H'\times W'},
\end{equation}
where $H'=H/P_H$, $W'=W/P_W$, and $E_{\mathrm{pos}}$ denotes the learnable positional embedding. Then we reshape $\mathbf{X}$ into
$B \times H' \times W' \times T \times D$
and employ a time aggregation layer~\cite{hao2024dpot} to integrate temporal information:
\begin{equation}
\mathbf{Y}_{b,h,w}
=
\sum_{t=1}^{T} \mathbf{X}_{b,h,w,t} \mathbf{W}_t,
\end{equation}
where $\mathbf{W}_t \in \mathbb{R}^{D \times D}$ is the learnable projection matrix for the $t$-th time step. The aggregated output is
$\mathbf{Y} \in \mathbb{R}^{B \times H' \times W' \times D}$.




\subsection{Spectral-Disentangled Neural Operator}
Spatiotemporal physical fields contain smooth global structures and sharp local variations. Therefore, we introduce a spectral-disentangled dual-branch architecture that transforms features into the frequency domain and separates low- and high-frequency components. The low-frequency branch captures shared dynamics, while the high-frequency branch models PDE-specific patterns with MoE.

\subsubsection{Low-Frequency Global Branch} After Fourier transformation of $\mathbf{Y}$ to extract low-frequency feature \(\mathbf{x}_{\mathrm{l}} \in \mathbb{R}^{B \times C \times H \times W}\), We employ AFNO~\cite{guibas2021adaptive} as the shared expert for low-frequency modeling. The input is first transformed into the frequency domain as \(\hat{\mathbf{x}}_{\mathrm{l}}=\mathcal{F}(\mathbf{x}_{\mathrm{l}})\). The channel dimension is divided into \(N\) blocks, and only the first \(K\) low-frequency modes are processed using a two-layer complex-valued MLP:
\begin{equation}
\mathbf{z}_{:K,:K}
=
\mathbf{W}_2\sigma\left(
\mathbf{W}_1\hat{\mathbf{x}}_{1_{:K,:K}}+\mathbf{b}_1
\right)+\mathbf{b}_2,
\end{equation}
where \(\mathbf{W}_1\) and \(\mathbf{W}_2\) are block-wise complex-valued weights, and \(\sigma\) denotes the activation function. The remaining frequency modes are set to zero, and the transformed features are mapped back to the spatial domain with a residual connection~\cite{he2016deep}, yielding \(\mathbf{x}_{\mathrm{lf}}=\mathcal{F}^{-1}(\mathbf{z})+\mathbf{x}_{\mathrm{l}}\).

\subsubsection{High-Frequency Local Branch}
To capture sharp local variations, such as small-scale vortices and shock boundaries, we apply a hard high-pass filter in the frequency domain. Given high-frequency feature \(\mathbf{x_h}\), \(\hat{\mathbf{x}}_{\mathrm{h}}=\mathcal{F}(\mathbf{x_h})\), a binary mask \(\mathbf{M}\in\{0,1\}^{H_f\times W_f}\) is defined as
\begin{equation}
\mathbf{M}(i,j)=
\begin{cases}
0, & i<K \ \text{and}\ j<K,\\
1, & \text{otherwise}.
\end{cases}
\end{equation}

The high-frequency component is then obtained as \(x_{\mathrm{hf}}=\mathcal{F}^{-1}(\hat{\mathbf{x}}\odot \mathbf{M})\) and fed into the MoE module to model fine-grained local structures and complex dynamics.

\subsubsection{Low-Rank Adapter-based Mixture of Experts}
High-frequency components, such as small-scale vortices and shock boundaries, are difficult to model with a single architecture. We therefore introduce a low-rank adapter-based MoE module that uses dynamic routing to enhance local high-frequency modeling. It contains 12 non-shared experts and one base expert, with the gate activating the Top-2 non-shared experts for each input.

\textbf{(1) Routing Strategy.} 
Since the MoE module processes high-frequency features, the router captures local variations by removing the spatial mean:
$\mathbf{x}_{\mathrm{hf}}=\mathbf{x}-\operatorname{Mean}_{H,W}(\mathbf{x})$.

A texture-aware routing representation is then constructed by combining channel activation and spatial variation:
\begin{equation}
\mathbf{g}_b
=
\left[
\operatorname{Mean}_{H,W}(\mathbf{x}_b);
\operatorname{Std}_{H,W}(\mathbf{x}_{\mathrm{hf},b})
\right]
\in\mathbb{R}^{2C}.
\end{equation}
This representation captures both channel responses and local fluctuations.

The routing probabilities are computed as \(\mathbf{p}_b=\operatorname{Softmax}(\mathbf{W}\mathbf{g}_b+\mathbf{b})\), where \(\mathbf{W}\in\mathbb{R}^{N\times 2C}\) and \(\mathbf{b}\in\mathbb{R}^{N}\) are learnable parameters. Following Top-2 routing~\cite{shazeer2017outrageously}, the two experts with the highest probabilities are selected. Let \(\mathcal{I}_b\) denote their index set. Their normalized routing weights are
\begin{equation}
\hat{p}_b^{(i)}
=
\alpha
\frac{p_b^{(i)}}
{\sum_{j\in\mathcal{I}_b}p_b^{(j)}},
\qquad i\in\mathcal{I}_b,
\end{equation}
where \(\alpha\) is the routing scale factor.

\textbf{(2) Depthwise Separable Convolution.}
To capture local details in high-frequency PDE features, we adopt depthwise separable convolution (DSC)~\cite{howard2017mobilenets} as the shared backbone of all experts. Given $\mathbf{x}\in\mathbb{R}^{B\times C\times H\times W}$, DSC applies depthwise convolution for spatial modeling and pointwise convolution for channel fusion, formulated as $\mathbf{z}_c=\mathbf{W}^{\mathrm{dw}}_c*\mathbf{x}_c$ and $\mathbf{u}=\mathbf{W}^{\mathrm{pw}}*\mathbf{z}$. Each convolution is followed by Group Normalization~\cite{wu2018group} and GELU~\cite{hendrycks2016gaussian}. With the residual connection, the DSC backbone is defined as
\begin{equation}
\mathcal{F}_{\mathrm{DSC}}(\mathbf{x})
=
\mathbf{x}
+
\operatorname{PW}(\operatorname{DW}(\mathbf{x})).
\end{equation}

\textbf{(3) Base Expert.}
Although PDEs exhibit distinct dynamics, their high-frequency components may share structures. We introduce a shared expert to learn transferable high-frequency representations across PDEs via parameter sharing~\cite{caruana1997multitask}.

The base expert adopts a residual depthwise separable convolutional block, formulated as $\mathbf{y}_{\mathrm{shared}}=\mathbf{x}+\mathcal{F}_{\mathrm{shared}}(\mathbf{x})$. It provides stable shared representations while reducing redundant learning among experts.

\textbf{(4) Specific Experts.}
The specific experts capture PDE-specific high-frequency patterns using a shared depthwise separable convolutional backbone and expert-specific low-rank adapters~\cite{hu2022lora}. For the $i$-th expert, the adapter consists of a down-projection $\mathbf{B}_i:C\rightarrow r$ and an up-projection $\mathbf{A}_i:r\rightarrow C$, formulated as
\begin{equation}
\mathbf{y}_i
=
\mathcal{F}_{\mathrm{base}}(\mathbf{x})
+
\mathbf{A}_i\!\left(\mathbf{B}_i(\mathbf{x})\right).
\end{equation}

The low-rank residual introduces expert-specific transformations with limited parameter overhead. The up-projection $\mathbf{A}_i$ is initialized to zero, making each non-shared expert initially equivalent to the shared backbone and improving training stability.

\textbf{(5) Routing and Feature Fusion.}
The shared expert provides general high-frequency representations across PDE systems, while the selected non-shared experts capture PDE-specific local patterns. For each input sample, the gating network selects the Top-2 non-shared experts. Let $\mathcal{I}_b$ denote the selected expert set for the $b$-th sample. The selected expert outputs are combined as
\begin{equation}
\mathbf{y}_{\mathrm{moe},b}
=
\sum_{i\in\mathcal{I}_b}
\hat{p}_b^{(i)}\mathbf{y}_{i,b},
\end{equation}
where $\hat{p}_b^{(i)}$ is the normalized routing weight.

The specialized and shared representations are fused using a learnable coefficient:
\begin{equation}
\mathbf{y}_{\mathrm{fusion}}
=
\lambda\mathbf{y}_{\mathrm{moe}}
+
(1-\lambda)\mathbf{y}_{\mathrm{shared}},
\;
\lambda=\operatorname{Sigmoid}(\beta).
\end{equation}

Finally, a global residual connection preserves the input feature information as $\mathbf{y}=\mathbf{x}+\mathbf{y}_{\mathrm{fusion}}$. This design combines shared and expert-specific representations for effective high-frequency PDE modeling.

\subsection{Knowledge-Guided Expert Aggregation}
\begin{figure}[t]
    \centering
    \includegraphics[width=\linewidth]{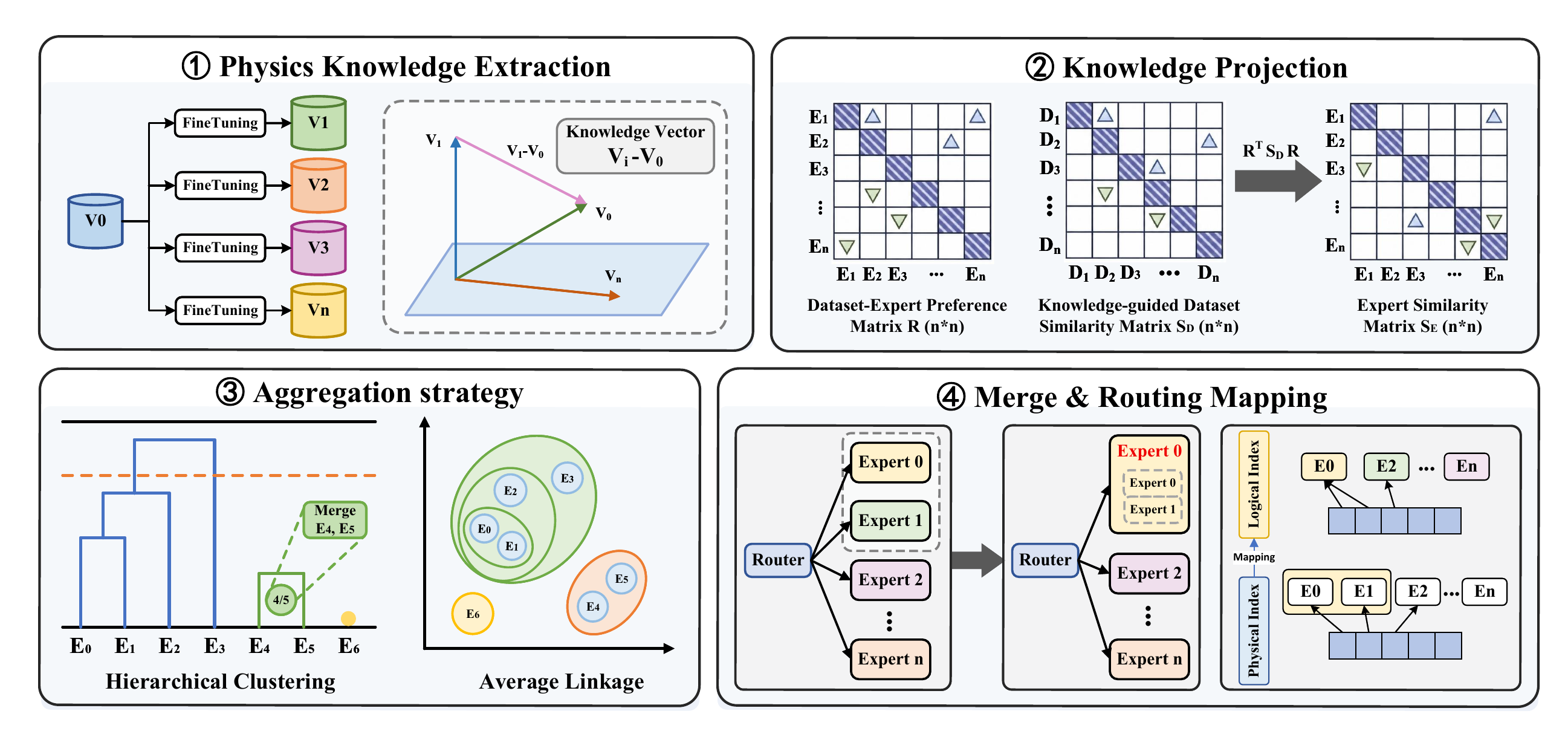}
    \caption{Knowledge-guided expert merging based on bilinear projection and hierarchical clustering.}
    \label{fig:mergeFigure}
\end{figure}

Then, we propose a knowledge-guided expert aggregation strategy that groups experts learning similar high-frequency representations, reducing parameter and memory overhead while preserving expert specialization, as shown in Fig. \ref{fig:mergeFigure}.

\subsubsection{Expert Similarity}
To measure expert similarity, we combine dataset-specific knowledge with expert activation preferences. Suppose the model is trained on \(D\) datasets with \(N\) non-shared experts. The average routing probabilities form a dataset--expert preference matrix \(\mathbf{R}\in\mathbb{R}^{D\times N}\).

Let \(\mathbf{v}_0\) denote the pretrained parameters and \(\mathbf{v}_d\) the parameters fine-tuned on dataset \(d\). The dataset-specific physical knowledge is represented by \(\Delta\mathbf{v}_d=\mathbf{v}_d-\mathbf{v}_0\). Their pairwise cosine similarities define
\begin{equation}
\mathbf{S}_D(d_1,d_2)
=
\frac{
\Delta\mathbf{v}_{d_1}^{\top}\Delta\mathbf{v}_{d_2}
}{
\|\Delta\mathbf{v}_{d_1}\|_2
\|\Delta\mathbf{v}_{d_2}\|_2
}.
\end{equation}

We then project the dataset-level similarities into the expert space:
\begin{equation}
\mathbf{S}_E^{\mathrm{raw}}
=
\mathbf{R}^{\top}\mathbf{S}_D \mathbf{R}
\in\mathbb{R}^{N\times N}.
\end{equation}
To reduce the effect of activation scale, we normalize it as
\begin{equation}
\mathbf{S}_E(i,j)
=
\frac{
\mathbf{S}_E^{\mathrm{raw}}(i,j)
}{
\sqrt{
\mathbf{S}_E^{\mathrm{raw}}(i,i)
\mathbf{S}_E^{\mathrm{raw}}(j,j)
}
+\varepsilon
},
\end{equation}
where $\varepsilon$ ensures stability. Larger $\mathbf{S}_E(i,j)$ indicates stronger knowledge similarity between experts $i$ and $j$.

\subsubsection{Expert Aggregation Strategy}
Based on \(S_E\), we apply threshold-constrained average-linkage hierarchical clustering~\cite{murtagh2012algorithms}. Each expert is initially treated as an individual cluster, and the similarity between clusters \(\mathcal{C}_a\) and \(\mathcal{C}_b\) is
\begin{equation}
\operatorname{Sim}(\mathcal{C}_a,\mathcal{C}_b)
=
\frac{1}{|\mathcal{C}_a||\mathcal{C}_b|}
\sum_{i\in\mathcal{C}_a}
\sum_{j\in\mathcal{C}_b}
\mathbf{S}_E(i,j).
\end{equation}
The most similar clusters are iteratively merged until their maximum similarity falls below a threshold \(\tau\).

Experts within each cluster are aggregated according to their global routing frequencies. For expert $i\in\mathcal{C}_k$, the normalized weight is $\omega_i=f_i/\sum_{j\in\mathcal{C}_k}f_j$, where $f_i$ denotes its global activation frequency. The aggregated adapter parameters are obtained as
\begin{equation}
\bar{\boldsymbol{\theta}}_k
=
\sum_{i\in\mathcal{C}_k}
\omega_i\boldsymbol{\theta}_i.
\end{equation}
The resulting aggregated experts replace the original experts, and the routing mapping is updated accordingly for inference.

\subsection{Loss Functions}
\subsubsection{Load Balancing Loss}
To promote balanced expert utilization, we adopt the load balancing loss~\cite{shazeer2017outrageously}:
\begin{equation}
\mathcal{L}_{\mathrm{aux}}
=
E\sum_{i=1}^{E}p_i f_i,
\end{equation}
where \(p_i\) and \(f_i\) denote the routing probability and activation frequency of expert \(i\), respectively.

\subsubsection{Main Task Loss}
We use the relative \(L_2\) reconstruction error (L2RE)~\cite{li2020fourier} as the main objective:
\begin{equation}
\mathcal{L}_{\mathrm{L2RE}}
=
\frac{
\left\|
\hat{\mathbf{u}}_i^{(c)}
-
\mathbf{u}_i^{(c)}
\right\|_2
}{
\left\|
\mathbf{u}_i^{(c)}
\right\|_2
}.
\end{equation}

\subsubsection{Orthogonality Loss}
To encourage expert diversity, we minimize the cosine similarity between the up-projection parameters of different non-shared experts:
\begin{equation}
\mathcal{L}_{\mathrm{orth}}
=
\frac{1}{|\Omega|}
\sum_{(i,j)\in\Omega}
\left(
\frac{
\mathbf{A}_i^{\top}\mathbf{A}_j
}{
\|\mathbf{A}_i\|_2
\|\mathbf{A}_j\|_2
}
\right)^2,
\end{equation}
where \(\Omega=\{(i,j)\mid i<j\}\).

\subsubsection{Overall Loss}
The overall objective is
\begin{equation}
\mathcal{L}_{\mathrm{total}}
=
\mathcal{L}_{\mathrm{L2RE}}
+
\lambda_1\mathcal{L}_{\mathrm{aux}}
+
\lambda_2\mathcal{L}_{\mathrm{orth}}.
\end{equation}
We set \(\lambda_1=0.001\). The coefficient \(\lambda_2\) decays from \(0.1\) to \(0.01\) over the first 400 epochs and reaches \(0\) at epoch 450.

\section{Experiments}

\subsection{Datasets and Baselines}
We evaluate our method on pre-training, fine-tuning, alternative data splits, unseen PDE transfer, model scaling, ablation studies, and expert merging.

\textbf{Datasets.}
We construct a mixed pre-training set of 12 PDE datasets with diverse equation types and parameter settings from FNO~\cite{li2020fourier}, PDEBench~\cite{takamoto2022pdebench}, PDEArena~\cite{gupta2022towards}, and CFDBench~\cite{luo2023cfdbench}.

\textbf{Baselines.}
We compare against representative neural operators and PDE foundation models, including FNO~\cite{li2020fourier}, UNet~\cite{ronneberger2015u}, GK-T~\cite{cao2021choose}, GNOT~\cite{hao2023gnot}, OFormer~\cite{li2022transformer}, MPP~\cite{mccabe2023multiple}, DPOT~\cite{hao2024dpot}, UniSolver~\cite{zhouunisolver}, MoE-POT~\cite{wang2026mixture}, NESTOR~\cite{sun2026nestor}, and Poseidon~\cite{herde2024poseidon}. UNet, GK-T, GNOT, and OFormer are trained separately on each dataset, while the others follow their original pre-training and transfer settings.


\begin{table*}[t]
\centering
\caption{The experiments are divided into four parts: standard small models, pre-training performance of foundation models, fine-tuning results, and model merging results. The evaluation metric is L2RE. The best result within each part is highlighted in \textbf{bold}, while the overall best result is emphasized in \colorbox[RGB]{196,216,242}{blue bold.}}
\label{tab:main_results}
\renewcommand{\arraystretch}{1.3}
\resizebox{\textwidth}{!}{%
\begin{tabular}{c|c|c|ccc|cccccccc|cc|c}
\toprule
\toprule
& L2RE & Activated & \multicolumn{3}{c|}{FNO-$\nu$}
& \multicolumn{8}{c|}{PDEBench CNS-($\eta,\zeta$), DR, SWE}
& \multicolumn{2}{c|}{PDEArena}
& CFDBench \\
& Model & Params
& 1e-5 & 1e-4 & 1e-3
& 1,0.1 & 1,0.01 & Avg.(1)
& 0.1,0.1 & 0.1,0.01 & Avg.(0.1)
& DR & SWE
& NS & NS-cond
& - \\
\midrule

\multirow{6}{*}{\rotatebox[origin=c]{90}{\textbf{Small model}}}
& FNO & 0.5M
& \textbf{0.116} & 0.0922 & 0.0156
& 0.151 & 0.108 & 0.130
& 0.230 & 0.076 & 0.153
& 0.0321 & 0.0091
& 0.210 & 0.384
& 0.0274 \\

& UNet & 25M
& 0.198 & 0.119 & 0.0245
& 0.334 & 0.291 & 0.313
& 0.569 & 0.357 & 0.463
& 0.0971 & 0.0521
& 0.102 & 0.337
& 0.209 \\

& FFNO & 1.3M
& 0.121 & 0.0503 & 0.0099
& 0.0212 & 0.0520 & 0.0366
& 0.162 & 0.0452 & 0.104
& 0.0571 & 0.0116
& 0.0839 & 0.602
& 0.0071 \\

& GK-T & 1.6M
& 0.134 & 0.0792 & \textbf{0.0098}
& 0.0341 & \textbf{0.0377} & \textbf{0.0359}
& 0.0274 & 0.0366 & 0.0320
& 0.0359 & 0.0069
& \textbf{0.0952} & 0.423
& 0.0105 \\

& GNOT & 1.8M
& 0.157 & \textbf{0.0443} & 0.0125
& \textbf{0.0325} & 0.0420 & 0.0373
& \textbf{0.0228} & 0.0341 & 0.0285
& 0.0311 & \textbf{0.0068}
& 0.172 & \textbf{0.325}
& \textbf{0.0088} \\

& OFormer & 1.9M
& 0.171 & 0.0645 & 0.0104
& 0.0417 & 0.0625 & 0.0521
& 0.0254 & \textbf{0.0205} & \textbf{0.0229}
& \textbf{0.0192} & 0.0072
& 0.135 & 0.332
& 0.0102 \\
\midrule

\multirow{5}{*}{\rotatebox[origin=c]{90}{\textbf{Pre-trained}}}
& MPP & 7M
& - & - & -
& - & - & 0.0442
& - & - & 0.0312
& \textbf{0.0168} & 0.0066
& - & -
& - \\

& DPOT & 7M
& 0.0976 & \textbf{0.0606} & 0.0095
& 0.0173 & 0.0397 & 0.0285
& 0.0132 & 0.0220 & 0.0176
& 0.0321 & 0.0056
& \textbf{0.125} & 0.384
& \textbf{0.0095} \\

& NESTOR & 13M
& 0.1195 & 0.0951 & 0.0093
& 0.0167 & 0.0373 & 0.0270
& 0.0120 & 0.0202 & 0.0161
& 0.0308 & 0.0052
& 0.132 & 0.409
& 0.0112 \\

& Unisolver & 12M
& \textbf{0.0600} & 0.0626 & 0.0148
& 0.0464 & 0.1172 & 0.0818
& 0.2510 & 0.1170 & 0.1840
& 0.7276 & 2.6851
& 0.1306 & 0.5649
& 0.0627 \\

& \textbf{Ours} & 11M
& 0.0960 & 0.0770 & \textbf{0.0087}
& \textbf{0.0120} & \textbf{0.0372} & \textbf{0.0246}
& \colorbox[RGB]{196,216,242}{\textbf{0.0116}} & \textbf{0.0184} & \textbf{0.0150}
& 0.0403 & \textbf{0.0048}
& 0.127 & \textbf{0.361}
& 0.0115 \\
\midrule

\multirow{3}{*}{\rotatebox[origin=c]{90}{\textbf{Fine-tune}}}


& DPOT-FT500 & 7M
& 0.0520 & 0.0367 & 0.0058
& 0.0112 & 0.0195 & 0.0153
& 0.0174 & 0.0138 & 0.0156
& 0.0148 & \colorbox[RGB]{196,216,242}{\textbf{0.0024}}
& 0.0910 & \textbf{0.280}
& 0.0039 \\

& NESTOR-FT500 & 13M
& \colorbox[RGB]{196,216,242}{\textbf{0.0505}}
& 0.0217
& 0.0043
& 0.0094
& \colorbox[RGB]{196,216,242}{\textbf{0.0134}}
& 0.01140
& \textbf{0.0123}
& 0.0083
& \textbf{0.0103}
& \colorbox[RGB]{196,216,242}{\textbf{0.0117}}
& 0.0026
& 0.0683
& 0.285
& \textbf{0.0038} \\

& Ours-FT500 & 11M
& 0.0543
& \textbf{0.0197}
& \colorbox[RGB]{196,216,242}{\textbf{0.0040}}
& \colorbox[RGB]{196,216,242}{\textbf{0.0090}}
& 0.0137
& \colorbox[RGB]{196,216,242}{\textbf{0.01135}}
& 0.0132
& \textbf{0.0082}
& 0.0107
& 0.0213
& 0.0045
& \colorbox[RGB]{196,216,242}{\textbf{0.0637}}
& 0.291
& \textbf{0.0038} \\
\midrule

\multirow{3}{*}{\rotatebox[origin=c]{90}{\textbf{Merge}}}
& Ours-merge & 11M
& \textbf{0.0541} & \colorbox[RGB]{196,216,242}{\textbf{0.0192}} & \colorbox[RGB]{196,216,242}{\textbf{0.0040}}
& \colorbox[RGB]{196,216,242}{\textbf{0.0090}} & \textbf{0.0137} & \colorbox[RGB]{196,216,242}{\textbf{0.01135}}
& \textbf{0.0127} & \colorbox[RGB]{196,216,242}{\textbf{0.0079}} & \colorbox[RGB]{196,216,242}{\textbf{0.0102}}
& \textbf{0.0197} & \textbf{0.0040}
& \colorbox[RGB]{196,216,242}{\textbf{0.0637}} & \colorbox[RGB]{196,216,242}{\textbf{0.279}}
& \colorbox[RGB]{196,216,242}{\textbf{0.0037}} \\

& Ours-OPCM & 11M
& 1.5940 & 1.2863 & 1.4523
& 5.0913 & 1.4027 & 3.2470
& 35.4204 & 4.1313 & 19.7657
& 0.8393 & 0.8393
& 1.4812 & 1.3098
& 0.7957 \\

& Ours-SMOE & 11M
& 0.0550 & 0.0200 & 0.0044
& 0.0094 & 0.0141 & 0.0118
& 0.0123 & 0.0084 & 0.0104
& 0.0120 & 0.0045
& 0.0640 & 0.2930
& 0.0037 \\
\bottomrule
\end{tabular}
}
\end{table*}

\subsection{Main Results}
Table \ref{tab:main_results} compares our method across pre-training, fine-tuning, and expert merging. Overall, our method achieves the best performance on 8 of the 12 tasks.

\textbf{Pre-training Stage.}
Our method achieves the best performance on 7 tasks, including 5 of the 6 PDEBench tasks, while maintaining low L2RE on most others. This demonstrates that the proposed MoE architecture effectively learns shared physical knowledge across heterogeneous PDEs with high parameter efficiency.

\textbf{Fine-tuning Stage.}
We fine-tune the pre-trained model for 500 epochs on each downstream dataset. The resulting model achieves the best performance on 6 tasks. Compared with training from scratch, pre-training enables faster adaptation and higher prediction accuracy, indicating strong transferability across PDE tasks.

\textbf{Expert Merging.}
We further compress the fine-tuned model by aggregating 12 non-shared experts into 6 according to their knowledge-guided similarities. Despite a 50\% reduction in expert count, Ours-Merge maintains or improves performance on most tasks. For example, L2RE on PDEArena NS-cond decreases from 0.291 to 0.281, demonstrating nearly lossless compression.

In contrast, OPCM~\cite{tang2026merging} suffers substantial degradation as the number of merged experts increases, with L2RE increasing by tens or even hundreds of times on some tasks. This suggests that generic merging methods struggle to preserve functional differences among experts. Our method instead identifies redundant experts using cross-task knowledge and routing preferences, reducing redundancy while preserving expert diversity and critical PDE knowledge.

Overall, the pre-training and fine-tuning paradigm improves performance and adaptation across PDE tasks, while knowledge-guided expert aggregation enables efficient compression with minimal performance degradation, demonstrating cross-task generalization and parameter efficiency.

\begin{table}[t]
\centering
\caption{Performance comparison of different methods under the new dataset split. The evaluation metric is L2RE. The best result within each column is highlighted in \textbf{bold}.}
\label{tab:new_dataset_split}
\renewcommand{\arraystretch}{1.2}
\resizebox{1\columnwidth}{!}{%
\begin{tabular}{l|c|cc|ccc|c}
\toprule
Model & Activated & \multicolumn{2}{c|}{FNO-$\nu$} & \multicolumn{3}{c|}{PDEBench} & CFDBench \\
L2RE & Params & 1e-5 & 1e-3 & 0.1,0.01 & SWE & DR & - \\
\midrule
FNO     & 10M & 0.1910 & 0.0245 & 0.0859 & 11.0000 & 0.5300 & 0.0060 \\
DPOT    & 7M  & 0.1070 & 0.0155 & 0.0168 & 0.0063 & 0.0577 & 0.0067 \\
MoE-POT & 17M & 0.0682 & 0.0077 & \textbf{0.0105} & 0.0064 & 0.0411 & \textbf{0.0053} \\
NESTOR  & 13M & 0.0674 & 0.0076 & 0.0159 & 0.0045 & 0.0184 & 0.0091 \\
Ours    & 10M & \textbf{0.0582} & \textbf{0.0051} & 0.0165 & \textbf{0.0024} & \textbf{0.0145} & 0.0056 \\
\bottomrule
\end{tabular}%
}
\end{table}

\begin{table}[t]
\centering
\caption{Comparison of generalization and transfer ability.}
\label{tab:Downstream}
\renewcommand\arraystretch{0.8}
\resizebox{1\linewidth}{!}{
\begin{tabular}{lccccc}
\toprule
Model & Activated Params & Wave-Layer & Wave-Gauss & Turbulence \\
\midrule
FNO       & 0.5M & - & - & 0.193 \\
MPP       & 7M   & - & - & 0.152 \\
DPOT      & 7M   & 0.0095 & 0.0075 & 0.135 \\
Poseidon  & 21M  & 0.29   & 0.29   & -     \\
MoE-POT   & 17M  & 0.05   & 0.06   & \textbf{0.0621} \\
\textbf{Ours} & 10M & \textbf{0.0085} & \textbf{0.0066} & 0.0662 \\
\bottomrule
\end{tabular}
}
\end{table}

\begin{figure}[t]
    \centering
    \includegraphics[width=1.0\linewidth]{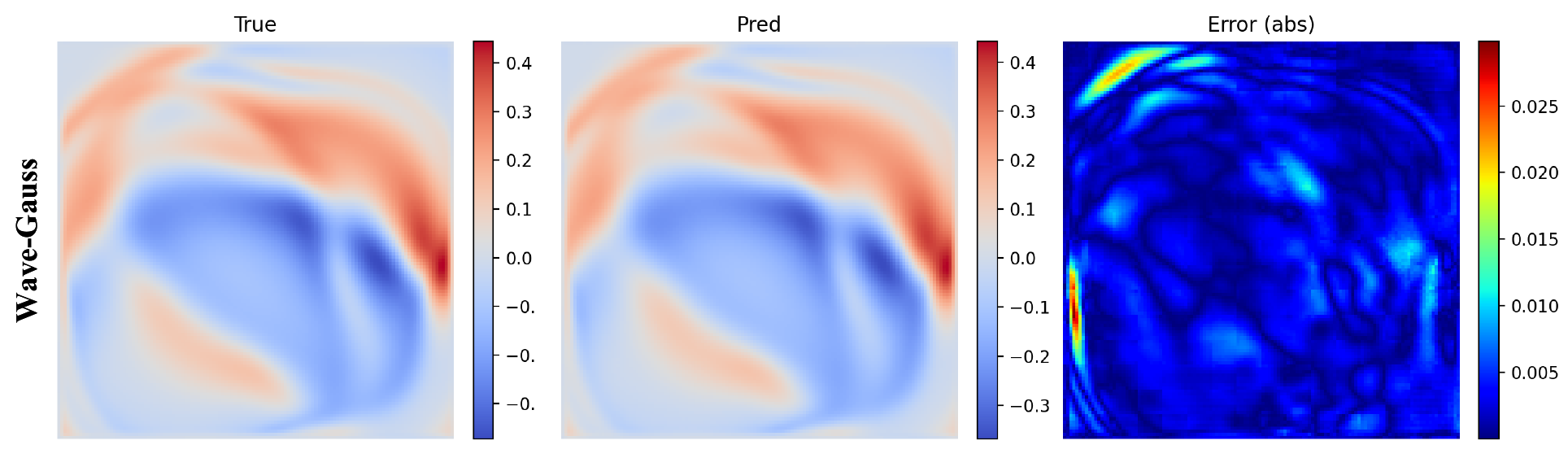}
    \caption{Visualization of the predicted physical field at the final time step for Wave-Gauss.}
    \label{fig:gauss}
\end{figure}

\begin{figure}[t]
    \centering
    \includegraphics[width=1\linewidth]{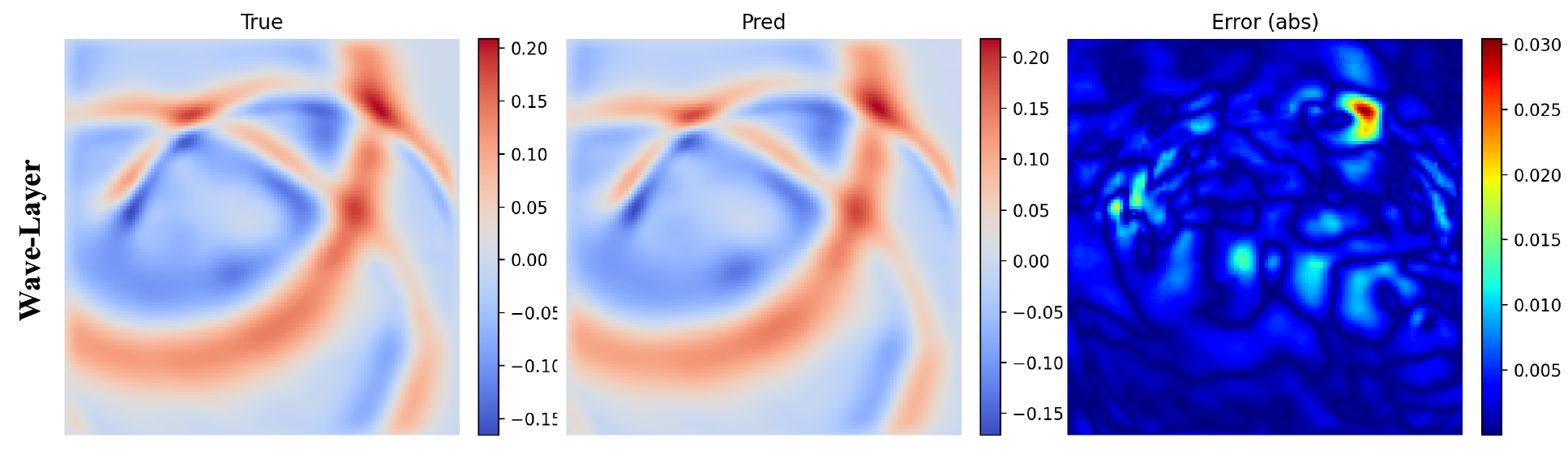}
    \caption{Visualization of the predicted physical field at the final time step for Wave-Layer.}
    \label{fig:layer}
\end{figure}

\subsection{Experiments on New PDE Dataset Splits}
To evaluate our model under different pre-training data configurations, we select six subsets from the original 12 PDE datasets and compare our method with state-of-the-art pre-trained models. The results are shown in Table \ref{tab:new_dataset_split}.

Our method achieves the best performance on four of the six datasets and remains competitive on the others, consistently outperforming pre-trained models on FNO-$\nu$, SWE, and DR. Even with the pre-training data size reduced by half, our method learns transferable physical representations from limited and heterogeneous PDE data, demonstrating strong downstream generalization.

\subsection{Downstream Task Experiments}
\begin{figure}[t]
    \centering
    \includegraphics[width=1\linewidth]{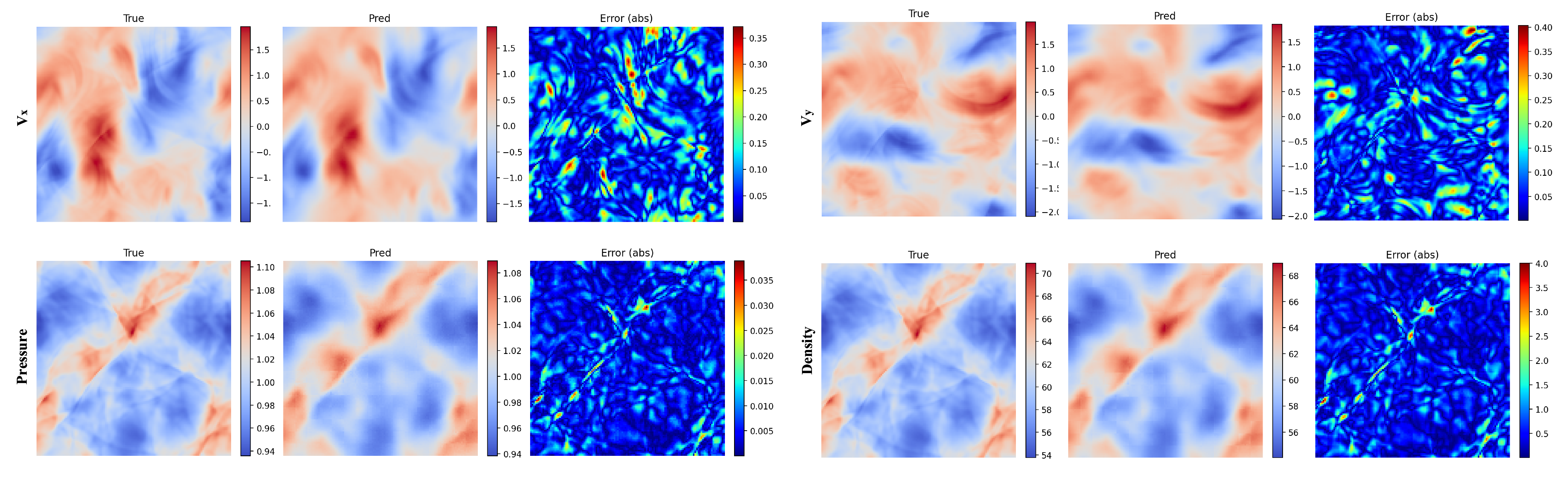}
    \caption{Visualization of the predicted physical variables at the final time step for 2D high-resolution turbulence.}
    \label{fig:turb}
\end{figure}

To evaluate generalization and transferability, we evaluate our model on a high-resolution 2D turbulence dataset and two unseen PDEs, Wave-Gauss and Wave-Layer. The results are reported in Table~\ref{tab:Downstream} and Figures~\ref{fig:gauss}--\ref{fig:turb}.

Our method achieves the best performance on both unseen PDEs. It obtains L2RE values of 0.0085 on Wave-Layer and 0.0066 on Wave-Gauss, achieving relative improvements of approximately 10.5\% and 12.0\% over the best baselines, respectively. It also achieves the second-best result on high-resolution turbulence and outperforms most baselines.

These results demonstrate that large-scale pre-training enables transferable representations and improves adaptation to unseen PDEs and complex downstream dynamics.

\subsection{Effectiveness Analysis of Expert Aggregation}

To evaluate the proposed aggregation strategy, we visualize the key matrices involved and compare downstream performance before and after aggregation.

\begin{figure}
    \centering
    \includegraphics[width=1\linewidth]{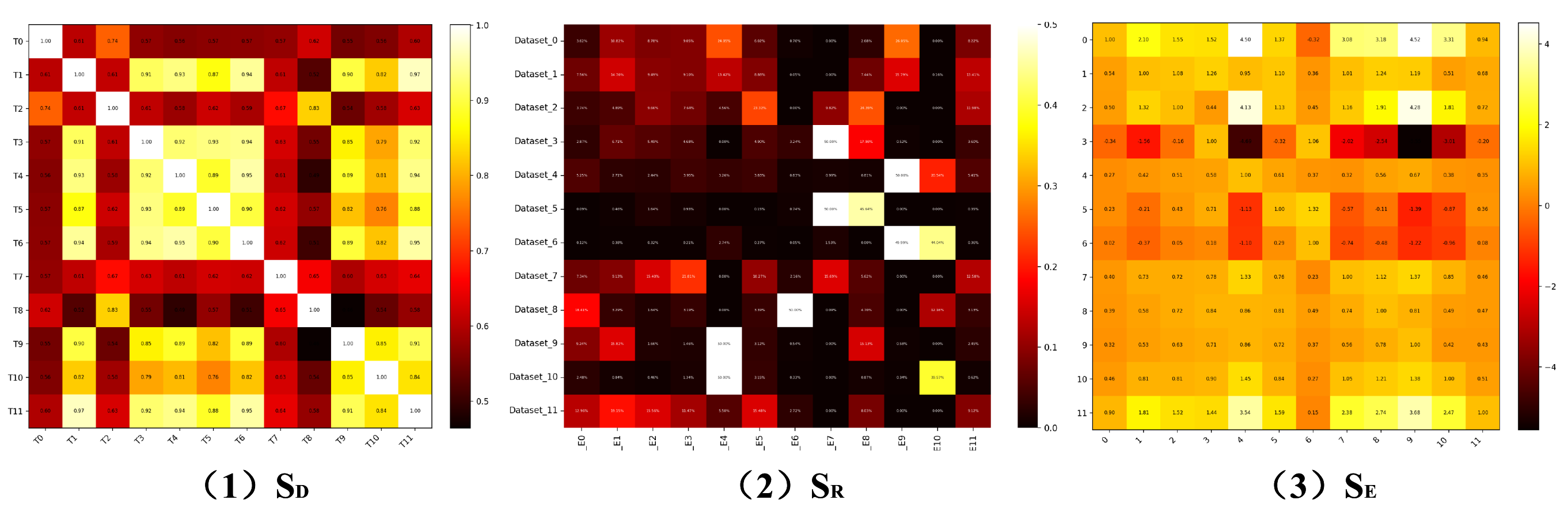}
    \caption{Visualization of the key matrices used in Knowledge-Guided expert merging: (1) dataset similarity matrix $S_D$, (2) dataset--expert preference matrix $R$, and (3) expert similarity matrix $S_E$.}
    \label{fig:matrix}
\end{figure}

Figure~\ref{fig:matrix}(1) shows the dataset similarity matrix $S_D$ , computed from $\Delta v_i=v_i-v_0$. Clear block structure is observed: datasets with similar physical mechanisms exhibit higher similarity, while different equation types show weaker correlations, indicating that the learned knowledge captures dataset-specific characteristics.
Figure~\ref{fig:matrix}(2) presents the dataset–expert preference matrix 
$R$. The sparse and locally concentrated pattern indicates that different datasets activate distinct experts, confirming that routing promotes expert specialization.
The knowledge-guided expert similarity matrix 
$S_E$, obtained by combining 
$S_D$ and 
$R$, is shown in Figure~\ref{fig:matrix}(3). High-similarity regions reflect knowledge overlap among experts, guiding the aggregation of redundant experts while preserving essential knowledge.
As shown in Table~\ref{tab:main_results}, reducing the number of experts from 12 to 6 achieves 50\% compression with negligible performance loss, and even improves performance on certain tasks (e.g., PDEArena NS-cond L2RE decreases from 0.291 to 0.281).

Overall, the strategy reduces parameter redundancy and improves representation consistency by merging experts with similar learned knowledge, balancing efficiency and generalization.


\begin{table}[t]
\centering
\caption{The impact of the number of non-shared experts.}
\label{tab:scaling}
\resizebox{1\linewidth}{!}{
\begin{tabular}{cccccccc} 
\toprule
Num. of experts & 1e-5 & 1e-3 & 0.1, 0.01 & SWE & DR & CFDBench & Avg. $\mathcal{L}_{2}$ \\
\midrule
4  & 0.0666 & 0.0054 & 0.0162 & 0.0029 & 0.0193 & 0.0079 & 0.0197 \\
8  & 0.0679 & 0.0049 & 0.0170 & 0.0029 & 0.0187 & 0.0083 & 0.0200 \\
12 & 0.0690 & 0.0050 & 0.0146 & 0.0026 & 0.0190 & 0.0076 & \textbf{0.0196} \\
16 & 0.0689 & 0.0048 & 0.0165 & 0.0040 & 0.0192 & 0.0077 & 0.0202 \\
20 & 0.0700 & 0.0049 & 0.0170 & 0.0028 & 0.0197 & 0.0092 & 0.0206 \\
24 & 0.0693 & 0.0049 & 0.0162 & 0.0031 & 0.0187 & 0.0082 & 0.0201 \\
\bottomrule
\end{tabular}
}
\end{table}

\subsection{Scaling Experiments}
We pre-train on six subsets of the 12 PDE datasets and vary the number of non-shared experts while fixing the number of activated experts. As shown in Table~\ref{tab:scaling}, 12 non-shared experts achieve the best performance.

Increasing the expert scale does not always improve performance, as excessive experts complicate routing and may weaken expert collaboration. Therefore, an appropriate expert scale is essential for MoE modeling.

\begin{table}[t]
\centering
\small 
\caption{Ablation experiments of our proposed model. “w/o” denotes the removal of the corresponding component.}
\label{tab:消融}
\resizebox{\columnwidth}{!}{%
\begin{tabular}{lcccccccc}
\toprule
Method & 1e-5 & 1e-3 & 0.1, 0.01 & SWE & DR & CFDBench & Avg. $\mathcal{L}_{2}$ & Promotion \\
\midrule
Ours & 0.0690 & 0.0050 & 0.0148 & 0.0027 & 0.0196 & 0.0077 & 0.0198 & - \\
w/o MoE & 0.0709 & 0.0054 & 0.0165 & 0.0031 & 0.0196 & 0.0079 & 0.0206 & 0.0008 \\
w/o High-Freq & 0.1070 & 0.0155 & 0.0168 & 0.0063 & 0.0577 & 0.0067 & 0.0350 & \textbf{0.0152} \\
w/o Aux Loss & 0.0694 & 0.0055 & 0.0154 & 0.0027 & 0.0185 & 0.0098 & 0.0202 & 0.0004 \\
w/o Orth Loss & 0.0692 & 0.0051 & 0.0168 & 0.0026 & 0.0189 & 0.0083 & 0.0202 & 0.0004 \\
\bottomrule
\end{tabular}}
\vspace{-6pt}
\end{table}

\subsection{Ablation Studies}
We conduct ablation studies on six PDEBench subtasks to evaluate the contribution of each key component in the proposed framework. The results are reported in Table~\ref{tab:消融}.

\textbf{High-Frequency Branch.}
Removing the high-frequency branch increases the average L2RE by 0.0152, causing the largest performance degradation among all variants. This confirms its critical importance in capturing fine-grained structures and sharp local variations.

\textbf{Mixture-of-Experts Architecture.}
Removing the MoE module increases the average L2RE by 0.0008, demonstrating that dynamic expert routing effectively improves the modeling of heterogeneous PDE patterns.

\textbf{Load-Balancing Loss.}
Removing the auxiliary load-balancing loss increases the average L2RE by 0.0004. Although the impact is relatively modest, it promotes balanced expert utilization and stabilizes training.

\textbf{Orthogonality Loss.}
Removing the orthogonality loss increases the average L2RE by 0.0004, indicating that it encourages diverse expert representations, reduces redundant feature learning, and promotes specialization among experts.

\section{Conclusion}
We presented SPEAR, a spectral-disentangled MoE neural operator with knowledge-guided expert aggregation for large-scale PDE pre-training. SPEAR separates shared low-frequency dynamics from PDE-specific high-frequency patterns and reduces expert redundancy through knowledge-guided aggregation while preserving performance. Extensive experiments demonstrate that SPEAR achieves strong pre-training, transferability, and generalization to unseen PDE systems, providing an efficient and scalable solution for PDE foundation models.

\bibliography{aaai2027}


\end{document}